\documentclass[pdflatex,sn-mathphys-num]{sn-jnl}

\usepackage{graphicx}%
\usepackage{multirow}%
\usepackage{amsmath,amssymb,amsfonts}%
\usepackage{amsthm}%
\usepackage{mathrsfs}%
\usepackage[title]{appendix}%
\usepackage{textcomp}%
\usepackage{manyfoot}%
\usepackage{booktabs}%
\usepackage{algorithm}%
\usepackage{algorithmicx}%
\usepackage{algpseudocode}%
\usepackage{listings}%

\usepackage[table]{xcolor}
\usepackage{colortbl}
\usepackage{placeins}

\usepackage{lineno}

\theoremstyle{thmstyleone}%
\theoremstyle{thmstyletwo}%

\theoremstyle{thmstylethree}%

\begin{document}

\title[Article Title]{Unlock Pose Diversity: Accurate and Efficient Implicit Keypoint-based Spatiotemporal Diffusion for Audio-driven Talking Portrait}


\author[1,4]{\fnm{Chaolong} \sur{Yang}}\email{Chaolong.Yang@liverpool.ac.uk}
\equalcont{These authors contributed equally to this work.}

\author[2]{\fnm{Kai} \sur{Yao}}\email{jiumo.yk@antgroup.com}
\equalcont{These authors contributed equally to this work.}

\author[3]{\fnm{Yuyao} \sur{Yan}}\email{yuyao.yan@xjtlu.edu.cn}

\author[6]{\fnm{Chenru} \sur{Jiang}}\email{Chenru.Jiang@dukekunshan.edu.cn}

\author[1,5]{\fnm{Weiguang} \sur{Zhao}}\email{Weiguang.Zhao@liverpool.ac.uk}

\author*[4]{\fnm{Jie} \sur{Sun}}\email{Jie.Sun@xjtlu.edu.cn}

\author[1]{\fnm{Guangliang} \sur{Cheng}}\email{Guangliang.Cheng@liverpool.ac.uk}

\author[7]{\fnm{Yifei} \sur{Zhang}}\email{yifei.zhang@cn.ricoh.com}

\author[6]{\fnm{Bin} \sur{Dong}}\email{bin.dong@dukekunshan.edu.cn}

\author*[6]{\fnm{Kaizhu} \sur{Huang}}\email{Kaizhu.Huang@dukekunshan.edu.cn}


\affil[1]{\orgdiv{Department of Computer Science}, \orgname{University of Liverpool}, \orgaddress{\city{Liverpool}, \postcode{L69 7ZX}, \country{UK}}}

\affil[2]{\orgname{Ant Group}, \orgaddress{\city{Hangzhou}, \postcode{
310000}, \country{China}}}

\affil[3]{\orgdiv{School of Robotic}, \orgname{Xi'an Jiaotong-Liverpool University}, \orgaddress{\city{Suzhou}, \postcode{215123}, \country{China}}}

\affil*[4]{\orgdiv{Department of Mechatronics and Robotics}, \orgname{Xi'an Jiaotong-Liverpool University}, \orgaddress{\city{Suzhou}, \postcode{215123}, \country{China}}}

\affil[5]{\orgdiv{Department of Foundational Mathematics}, \orgname{Xi'an Jiaotong-Liverpool University}, \orgaddress{\city{Suzhou}, \postcode{215123}, \country{China}}}

\affil*[6]{\orgdiv{Digital Innovation Research Center}, \orgname{Duke Kunshan University}, \orgaddress{\city{Kunshan}, \postcode{215316}, \country{China}}}

\affil[7]{\orgname{Ricoh Software Research Center}, \orgaddress{\city{Beijing}, \postcode{100027}, \country{China}}}


\abstract{Audio-driven single-image talking portrait generation plays a crucial role in virtual reality, digital human creation, and filmmaking. Existing approaches are generally categorized into keypoint-based and image-based methods. Keypoint-based methods effectively preserve character identity but struggle to capture fine facial details due to the fixed points limitation of the 3D Morphable Model. Moreover, traditional generative networks face challenges in establishing causality between audio and keypoints on limited datasets, resulting in low pose diversity. In contrast, image-based approaches produce high-quality portraits with diverse details using the diffusion network but incur identity distortion and expensive computational costs. In this work, we propose KDTalker, the first framework to combine unsupervised implicit 3D keypoint with a spatiotemporal diffusion model. Leveraging unsupervised implicit 3D keypoints, KDTalker adapts facial information densities, allowing the diffusion process to model diverse head poses and capture fine facial details flexibly. The custom-designed spatiotemporal attention mechanism ensures accurate lip synchronization, producing temporally consistent, high-quality animations while enhancing computational efficiency. Experimental results demonstrate that KDTalker achieves state-of-the-art performance regarding lip synchronization accuracy, head pose diversity, and execution efficiency. Our codes are available at \url{https://github.com/chaolongy/KDTalker}.}




\maketitle
\section{Introduction}\label{sec1}
The generation of natural and dynamic talking portraits has wide-ranging applications in computer vision and digital content creation, such as virtual reality, digital human creation, and film and television production. Creating realistic talking head videos requires precise lip synchronization and natural pose variations, which has been a challenging task. Early research~\cite{suwajanakorn2017synthesizing, kr2019towards, prajwal2020lip, cheng2022videoretalking} primarily focused on lip synchronization, but by neglecting the diversity of head movements and facial expressions, these methods often produced stiff animations that lacked realism. As technology has advanced, researchers have increasingly recognized that relying solely on lip movements is insufficient to meet the demand for realism, leading to more comprehensive exploration of facial animation, including head poses, eye movements, and facial expressions.

\begin{figure}[ht]
   \centering
   \includegraphics[width=0.95\linewidth]{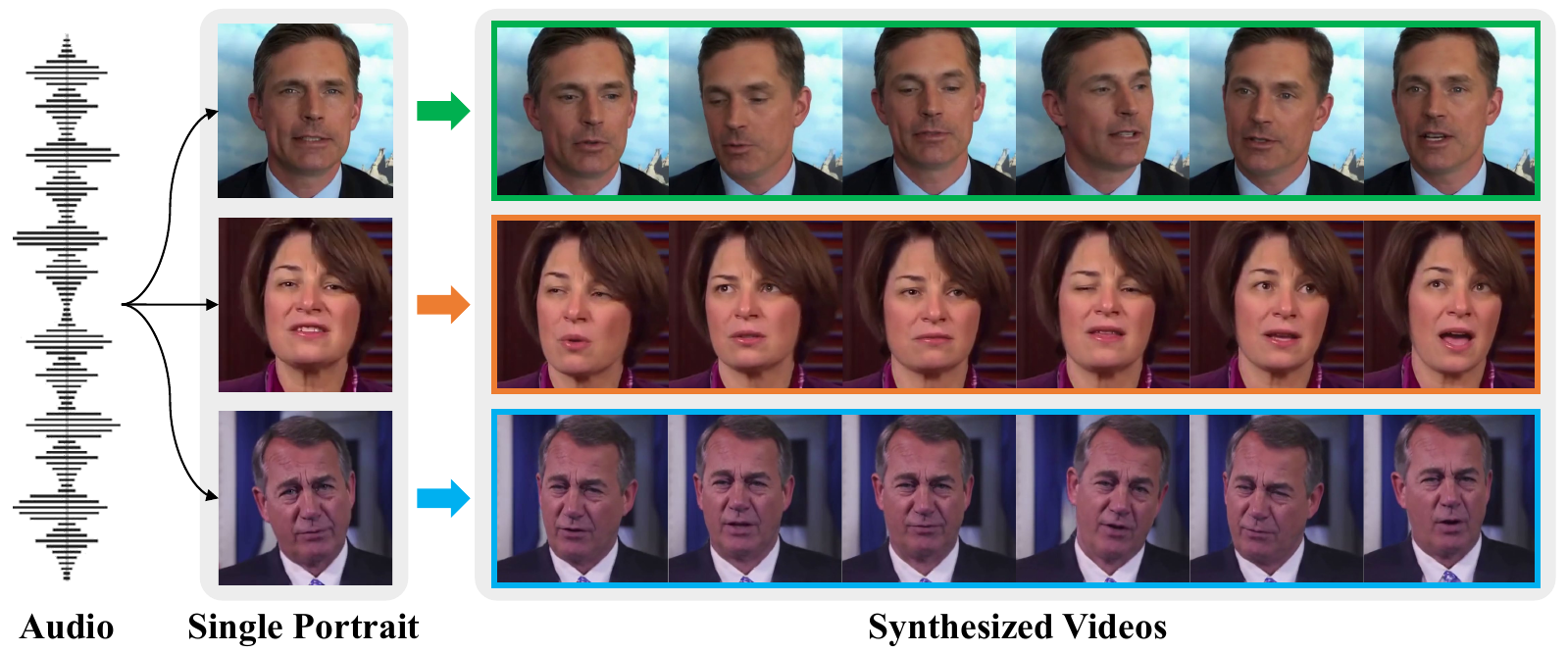}
   \caption{The proposed KDTalker:  A keypoint-based spatiotemporal diffusion framework that generates synchronized, high-fidelity talking videos from audio and a single image, enhancing pose diversity and expression detail with realistic, temporally consistent animations.}
   \label{fig:Ours_results}
\end{figure}

Previous approaches can be broadly categorized into keypoint-based and image-based methods to address these challenges. Explicit keypoint-based methods~\cite{zhang2023sadtalker, yereal3d, gururani2023space}, such as SadTalker~\cite{zhang2023sadtalker}, rely on audio to predict a 3D Morphable Model (3DMM)~\cite{blanz19993dmm} representation, which is mapped to unsupervised implicit 3D keypoints and input to a pre-trained facial rendering to generate facial animations that preserve character identity. Image-based methods~\cite{shen2023difftalk, liu2024anitalker, wei2024aniportrait} generate talking faces by directly training an image generator, typically producing high-quality images thanks to powerful pre-trained models such as Stable Diffusion~\cite{rombach2022high}.

However, keypoint-based methods struggle to capture subtle facial movements, such as eye movement and frowning, due to the fixed nature of traditional 3DMM keypoints, limiting the expression detail of the generated animations. Additionally, these methods use conventional generative networks like Variational Autoencoders (VAEs)~\cite{kingma2013vae} and Generative Adversarial Networks (GANs)~\cite{goodfellow2020gan}, which fail to effectively model the causal relationship between audio and key points on limited datasets, leading to poor diversity. On the other hand, image-based methods demand significant computational resources, and the image diffusion process considerably slows down generation speed, hindering real-time application. Moreover, while these methods excel in detail, they often suffer from identity distortion, leading to a loss of character features in the generated faces. Notably, even latent-space approaches like VASA-1~\cite{xu2024vasa} sacrifice explicit head pose control for implicit motion modeling, limiting motion flexibility.

To generate more nuanced facial expressions and diverse head poses while maintaining efficiency, we propose a novel implicit keypoint-based spatiotemporal diffusion framework, KDTalker. This model uniquely combines the flexibility of unsupervised implicit 3D keypoint-driven methods with the diversity capabilities of diffusion models. KDTalker predicts unsupervised implicit 3D deformation keypoints and transformation parameters from audio, capturing variations in lip movements, facial expressions, and head poses. These transformed keypoints are used for facial rendering to match the dynamics of the audio sequence. Unlike traditional 3DMM, our model does not rely on fixed keypoint positions. Instead, the learned keypoints can adapt to varying facial information densities, allowing for a flexible capture of subtle motion details through the diffusion process, significantly enhancing the expressiveness of the generated results. Additionally, KDTalker incorporates a spatiotemporal attention mechanism to capture long-range dependencies between audio and 3D keypoint mappings, ensuring that the generated animations are not only natural and smooth but also precisely synchronized with the audio.

\begin{figure}[t]
  \centering
   \includegraphics[width=.70\linewidth]{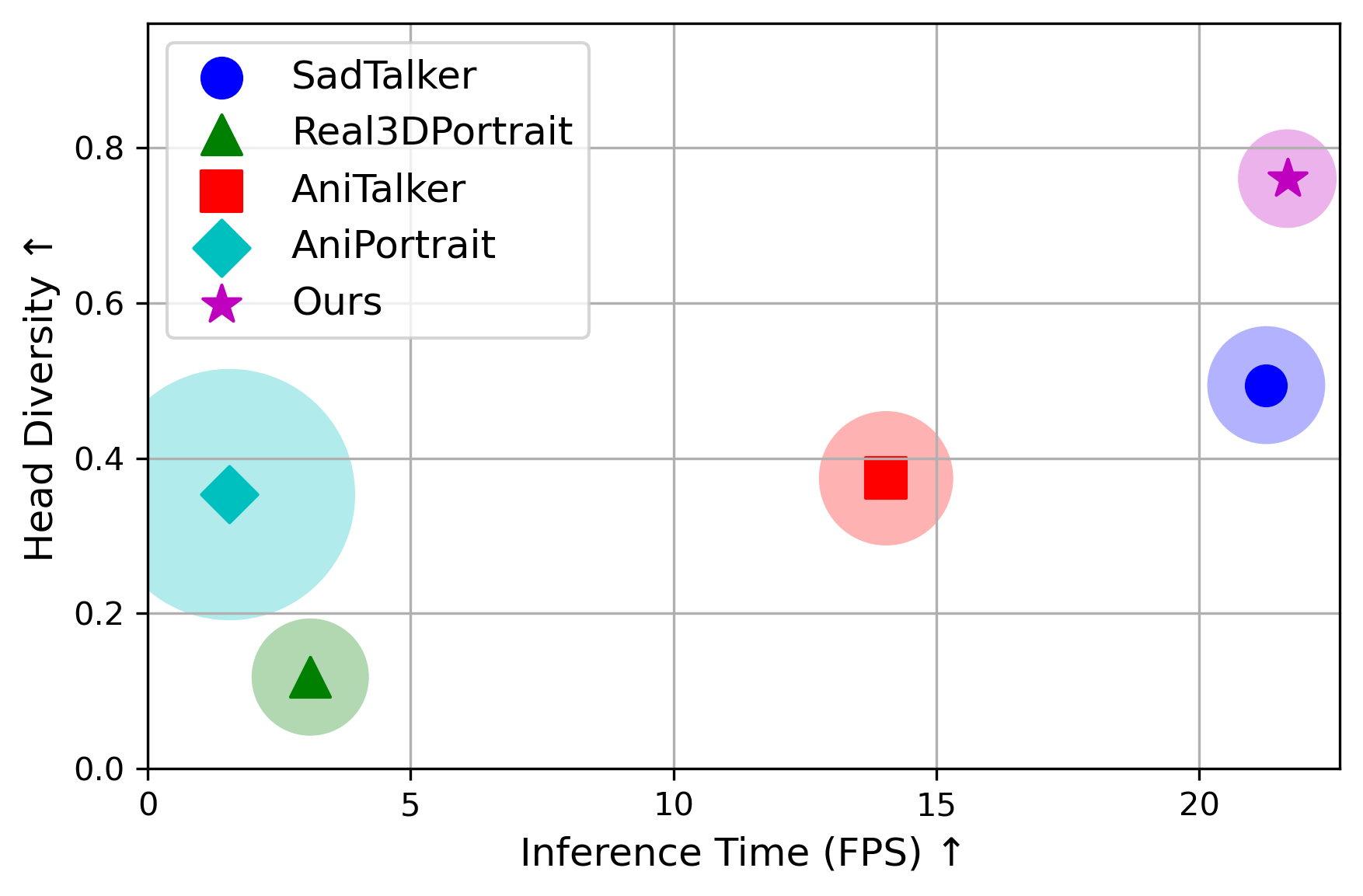}
   \caption{Inference Time vs Head Diversity $\&$ LSE-D. The value of LSE-D (Lip Sync Error Distance), a metric quantifying the alignment between lip movements and audio, is represented by the size of the circle. A smaller circle indicates a lower LSE-D value, reflecting better lip sync performance.}
   \label{fig:Inference Time vs Head Diversity & LSE-D}
\end{figure}

In summary, our KDTalker framework differs from previous works in three key aspects. First, unlike explicit keypoint-based methods~\cite{zhang2023sadtalker, yereal3d, gururani2023space} that rely on fixed 3DMM keypoints or landmarks, our unsupervised implicit keypoints dynamically adapt to facial feature densities, enabling flexible motion capture via diffusion modeling. This removes the rigidity of predefined keypoints and simplifies processing. Second, while conventional keypoint-driven approaches use VAEs/GANs~\cite{kingma2013vae, goodfellow2020gan}, our spatiotemporal diffusion model enhances expressiveness by modeling causal audio-to-keypoint relationships. Third, compared to non-keypoint image-based methods~\cite{shen2023difftalk, liu2024anitalker, wei2024aniportrait}, which suffer from high computational costs and identity distortion, KDTalker achieves efficient, real-time generation while preserving identity through keypoint-guided rendering. Additionally, our framework offers explicit control over head poses, overcoming the motion constraints of latent-space methods like VASA-1~\cite{xu2024vasa}. This integration of implicit keypoints and diffusion modelling achieves higher lip synchronization accuracy, richer head poses diversity and more efficient generation speed, as shown in Fig.~\ref{fig:Inference Time vs Head Diversity & LSE-D}. The results of pose diversity in the video generated by KDTalker can be seen in Fig.~\ref{fig:Ours_results}. Our contribution is threefold:

\begin{itemize}
    \item We propose a novel implicit keypoint-diffusion-based audio-driven talking portrait framework, named KDTalker, that utilizes spatiotemporal-aware attention to capture long-term dependencies between keypoints and audio, allowing for more accurate and contextually aligned facial movements.
    \item We are the first to integrate unsupervised implicit 3D keypoint-driven methods with diffusion models, significantly improving head pose and facial expression diversity while maintaining efficient computational performance suitable for real-time applications.
    \item Experimental results demonstrate that KDTalker achieves state-of-the-art performance in lip synchronization accuracy, head poses diversity and generation speed.
\end{itemize}


\section{Related Work}\label{sec2}
\subsection{Audio-driven One-shot Talking Portrait}
Early research~\cite{suwajanakorn2017synthesizing, kr2019towards, cheng2022videoretalking, prajwal2020lip} primarily focused on generating accurate lip-synced talking portrait videos from audio. However, these methods often had limitations, leading to fixed head movements that concentrated mainly on mouth motion, restricting overall realism. With the advent of deep learning, attention shifted to more comprehensive models that could capture not only mouth movements but also natural head poses and facial expressions, providing a more realistic depiction. Current approaches can be broadly divided into keypoint-based~\cite{chen2019hierarchical, zhou2020makelttalk, ren2021pirenderer, zhang2021flow, zhang2023sadtalker} and image-based~\cite{shen2023difftalk, ma2023dreamtalk, liu2024anitalker, wei2024aniportrait, kim2024moditalker, tian2024emo, xu2024vasa} methods.

Previous keypoint-based methods primarily rely on supervised keypoints, such as 3DMM~\cite{ren2021pirenderer, zhang2021flow, zhang2023sadtalker} or facial landmarks~\cite{chen2019hierarchical, zhou2020makelttalk, gururani2023space}, which map audio features to predefined keypoints for lips, blinks, and head poses. 
Although 3DMM provides a structured 3D parameterization, its fixed keypoint configuration constrains the capture of fine facial details and nuanced expressions. 
For instance, SadTalker~\cite{zhang2023sadtalker} employs unsupervised implicit 3D keypoints in the final rendering stage while it is fundamentally based on initial 3DMM mappings, thus inheriting the limitations of supervised 3DMM models. Furthermore, SPACE~\cite{gururani2023space} rely on predefined explicit keypoints as intermediate representations, limiting their flexibility. 
In contrast, KDTalker directly predicts implicit keypoints from audio, significantly simplifying intermediate processing. The distances and distribution relationships among the unsupervised implicit 3D keypoints dynamically adapt to facial feature density, enabling flexible control of expression details and head motion amplitude.

Image-based approaches~\cite{shen2023difftalk, liu2024anitalker, wei2024aniportrait} typically rely on pre-trained image generators, such as Stable Diffusion~\cite{rombach2022high}, to directly generate facial frames, often yielding high-quality images with detailed textures. Further research has focused on enhancing generation quality~\cite{ma2023dreamtalk, kim2024moditalker, tian2024emo}.
However, these methods are computationally demanding and require considerable inference time due to the image generation process, making real-time application challenging. Additionally, while these methods produce high visual fidelity, they tend to suffer from identity distortion, causing a loss of consistency in character features across frames. 
Specifically, VASA-1~\cite{xu2024vasa} adopts an implicit latent space to model audio-driven motion, which restricts direct control over the generated character’s head pose and movement amplitude. In contrast, our KDTalker framework leverages transformation parameters, allowing for more control over head posture, thereby offering greater flexibility in pose adjustment.

\subsection{Video-driven One-shot Talking Portrait}
Video-driven one-shot talking portrait synthesis involves transferring the motion from a driving video to a target portrait, aiming to reproduce the head movements, expressions, and lip-sync from the source in the target. Video-driven methods focus on motion transfer within the same domain (video-to-video), making the task relatively easier compared to audio-driven methods. Researchers have extensively explored this field by learning intermediate motion representations, such as unsupervised landmarks~\cite{siarohin2019first, wang2021facevid2vid, hong2022depth, wang2022one, zhao2022thin, guo2024liveportrait}, 3DMM~\cite{ren2021pirenderer, doukas2021headgan, yin2022styleheat}, and latent animations~\cite{wang2022latent, mallya2022implicit}. Our approach, KDTalker, leverages the latest video-driven method, LivePortrait~\cite{guo2024liveportrait}, to generate implicit unsupervised 3D keypoints. We then train a keypoint-based spatiotemporal diffusion model using these keypoints. During inference, KDTalker predicts implicit unsupervised 3D keypoints from the input reference image and audio, fed into LivePortrait for rendering and video generation. This approach effectively bridges the gap between video-driven and audio-driven tasks, ensuring the natural synchronization of facial movements with the audio.

\section{Method}\label{sec3}
\begin{figure*}[t]
\setlength {\abovecaptionskip} {0.cm}
\setlength {\belowcaptionskip} {0.cm}
  \centering
   \includegraphics[width=1.0\linewidth]{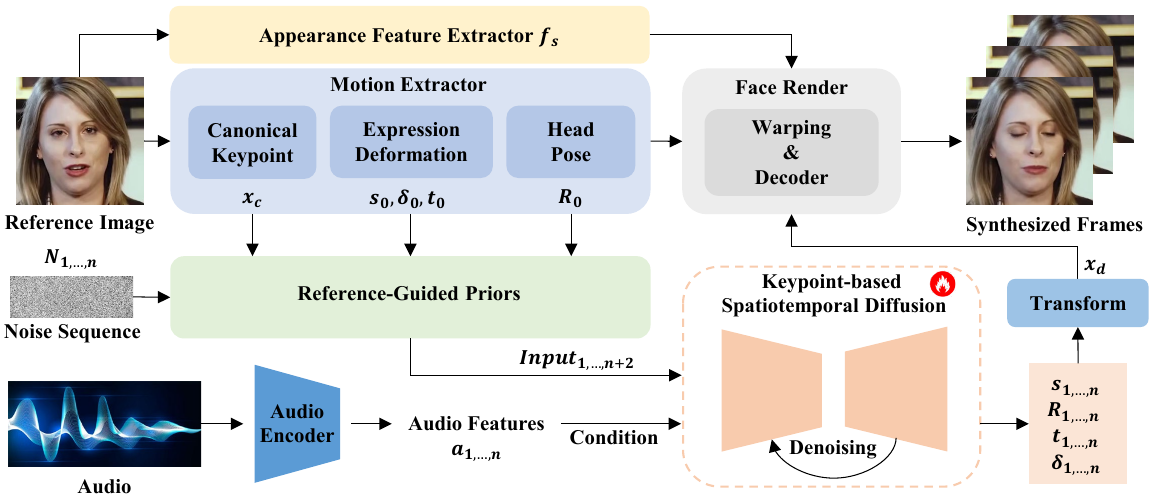}
   \caption{Overview of the proposed KDTalker for talking portrait synthesis.}
   \label{fig:overview}
\end{figure*}

\subsection{Overview}
As illustrated in Fig.~\ref{fig:overview}, our framework generates talking head videos by using unsupervised 3D facial keypoints as an intermediate representation. The process begins with the motion extractor extracting motion information from the reference image, while the audio encoder extracts audio features to serve as conditions for the diffusion model. Next, the Reference-Guided Priors module integrates this motion information with a noise sequence, preparing the input for the model. The Keypoint-based Spatiotemporal Diffusion module then uses this input and audio feature to predict expression deformation keypoints $\delta$, scaling and translation parameters $s$ and $t$, as well as head rotation $R$. Finally, the predicted values ($s$, $R$, $t$, $\delta$) are transformed into driving keypoints $x_{d}$, which are passed to the Face Render module. Through warping and decoding, this module generates $N$ video frames, producing a lifelike, synchronized animation that aligns with the input audio. Each component is detailed further in the following sections: audio and image processing in Sec.~\ref{sec:image and audio processing}, reference-guided priors in Sec.~\ref{sec:reference-guided priors}, keypoint-based spatiotemporal diffusion in Sec.~\ref{sec:keypoint-based spatiotemporal diffusion},  and keypoint-driven image animator in Sec.~\ref{sec:face render}

\subsection{Image and Audio Processing}
\label{sec:image and audio processing}
The Motion Extractor module leverages the pre-trained LivePortrait~\cite{guo2024liveportrait} framework to derive critical elements from the reference image. These include canonical keypoints that outline the fundamental facial structure, expression deformation keypoints capturing nuanced facial movements driven by speech and emotion, and head pose parameters that denote head orientation. Through an unsupervised learning process, LivePortrait generates flexible 3D keypoints without fixed positions, allowing canonical keypoints to adapt across different reference images. This flexibility enables the model to capture fine-grained expressions and a wide range of head poses with greater accuracy. The Appearance Feature Extractor extracts the appearance features of the reference image $f_{s}$, preserving the visual identity of the subject in the Face Render process. Together, these components ensure identity retention and structural coherence in the generated videos. Concurrently, the audio feature extraction module employs the pre-trained Wav2Lip~\cite{prajwal2020lip} AudioEncoder to capture audio-derived features, which serve as conditioning inputs for the model.

\subsection{Reference-Guided Priors}
\label{sec:reference-guided priors}
\begin{figure}[t]
\setlength {\abovecaptionskip} {0.cm}
\setlength {\belowcaptionskip} {0.cm}
   \centering
   \includegraphics[width=0.70\linewidth]{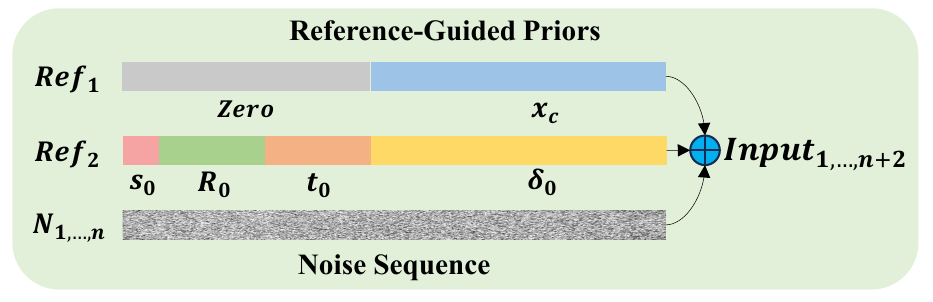}
   \caption{Reference-Guided Priors.}
   \label{fig:Reference-Guided Priors}
\end{figure}

The Reference-Guided Priors module integrates motion information from two key reference frames with a noise sequence to guide the generation process, as shown in Fig.~\ref{fig:Reference-Guided Priors}. Initially, the canonical keypoints $x_{c}$ extracted from the reference image are used as the input for the first frame, with zero-padding applied to align other parameters. This setup provides the model with prior information on the facial structure, ensuring the canonical keypoints serve as a foundation for subsequent frames. In our ablation study, we demonstrate the importance of this prior to achieving accurate lip synchronization. Specifically, the expression deformation $\delta_{0}$ aligns with $x_{c}$, while other parameters are aligned with zero-padding. These priors ($Ref_{1}$ \& $Ref_{2}$) are then integrated with a noise sequence $N_{1, \ldots, n}$ to construct the $Input_{1, \ldots, n+2}$ for the diffusion model. By initializing the process with reference frames, the model conditions the noise sequence on structured motion information, enabling temporally coherent motion generation that aligns with the original facial structure and adapts to the audio-driven dynamics.

\subsection{Keypoint-based Spatiotemporal Diffusion}
\label{sec:keypoint-based spatiotemporal diffusion}
The Keypoint-based Spatiotemporal Diffusion module serves as the core generative component, leveraging latent diffusion to predict motion parameters that create realistic, temporally consistent talking head animations. Given the structured priors and audio conditioning, the task can be formally defined as learning a mapping $f: (x_{c}, s_{0}, R_{0}, t_{0}, \delta_{0}, N_{1, \ldots, n}, a_{1, \ldots, n}) \rightarrow $ $(s, R, t, \delta)_{1, \ldots, n}$. These inputs are iteratively denoised by the diffusion model to obtain $N$-dimensional expression deformation keypoints and transformation parameters. Subsequently, the canonical keypoints $x_{c}$ are transformed according to the following equation:
\begin{equation}
    x_{d} = s \cdot (x_{c} \cdot R + \delta) + t,
  \label{eq:1}
\end{equation}
where $x_{d}$ represents the deformed keypoints.

\subsubsection{Diffusion Process}
The forward diffusion process begins by initializing a latent variable $z_{0}$ which encapsulates the facial features, including deformation keypoints and transformation parameters. The model then introduces noise progressively over $T$ steps to corrupt this latent representation. This process can be mathematically expressed as:
\begin{equation}
    \begin{aligned}
        z_{t} &= \sqrt{1 - \beta_{t}} z_{t-1} + \sqrt{\beta_{t}} \epsilon_{t-1},\\
          &= \sqrt{\bar{\alpha_{t}}} z_{0} + \sqrt{1 - \bar{\alpha_{t}}} \epsilon,
    \end{aligned}
  \label{eq:2}
\end{equation}
where $\epsilon \sim  N(0, I)$, $\beta_{t}$ is the diffusion rate corresponding to the $t$-th step in the increasing sequence with a value range between (0, 1), and $\bar{\alpha_{t}}=\prod_{i=1}^{t}(1-\beta_{i})$ is a parameter expression defined based on the diffusion rate $\beta_{t}$.

In the context of our framework, the forward process is conditioned on both the reference image and audio features. During training, the reference-guided prior module adds noise to the real keypoints and transformation parameters, and combines the motion information of the reference image to form structured input. These structured input, along with audio features, are then processed through the model to predict the added noise.

The goal of training the diffusion model is to learn a reverse process that reconstructs the original latent representation $z_{0}$ from the noisy latent variables $z_{t}$. This reverse process is modelled by a neural network $\epsilon_{\theta}$, which attempts to predict the noise added at each timestep $t$. In particular, $\epsilon_{\theta}$ here refers to our spatiotemporal-aware attention network. The corresponding loss function for training is the mean squared error (MSE) between the predicted noise and the actual noise $\epsilon_{t}$:
\begin{equation}
    \pounds (\theta) = \mathbb{E}_{t, z_{t}, a, \theta} [\left\| \epsilon_{t} - \epsilon_{\theta}(z_{t}, a, t) \right\|_{2}^{2}],
  \label{eq:3}
\end{equation}
where $\epsilon_{\theta}(z_{t}, a, t)$ is the model's prediction of the noise added at time $t$, $a$ denotes the extracted audio features.

During inference, starting from a Gaussian noise $z_{t}$, the model iteratively applies the reverse diffusion process to denoise $z_{t}$ and recover the original latent variable $z_{0}$. In our work, $z_{0}$ refers to deformation keypoints $\delta$, and transformation parameters ($s$, $R$, $t$). This reverse process can be described as:
\begin{equation}
    z_{t-1} = z_{t} - \beta_{t} \epsilon_{\theta}(z_{t}, a, t) + \sigma_{t} \epsilon,
  \label{eq:4}
\end{equation}
where $\sigma_{t}$ adjusts the noise variance at each step, controlling the amount of added noise and $\epsilon$ is sampled from a Gaussian distribution, introducing random variability into the process.

\subsubsection{Spatiotemporal-Aware Attention Network}
The Spatiotemporal-Aware Attention Network is designed to maintain both spatial coherence in facial regions and temporal consistency in motion, ensuring that keypoints evolve smoothly across frames while synchronizing with audio input. As illustrated in Fig.~\ref{fig:Spatiotemporal-Aware Attention Network}, this attention network integrates spatial and temporal information directly within the diffusion process to achieve coordinated and realistic motion of facial keypoints. The network consists of several key components:

\begin{figure}[t]
  \centering
   \includegraphics[width=0.65\linewidth]{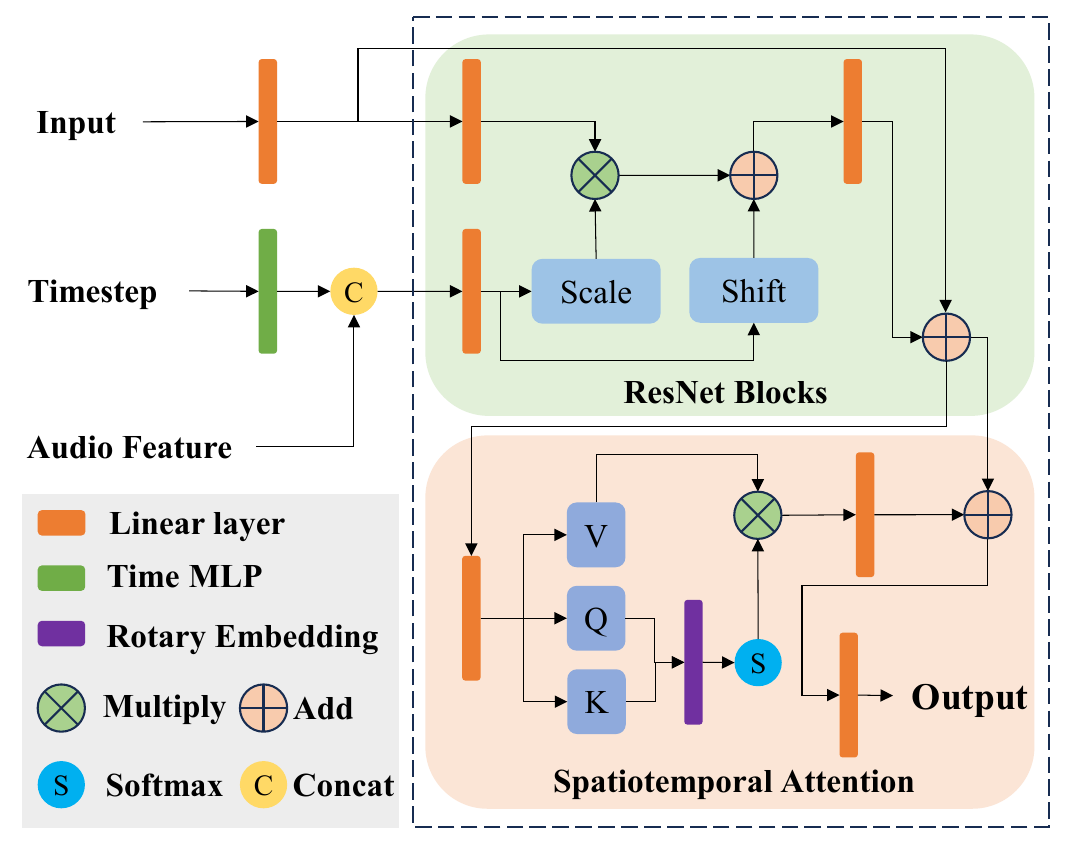}
   \caption{Spatiotemporal-Aware Attention Network.}
   \label{fig:Spatiotemporal-Aware Attention Network}
\end{figure}

\bmhead{Timestep and Audio Feature Integration} 
In each diffusion step, the sampling timestep is encoded by a Time MLP module to generate a temporal embedding vector, which is combined with audio features. This combined vector serves as a conditional input to the diffusion model, facilitating audio-driven learning that captures both temporal progression and spatial variations of facial keypoints.

\bmhead{ResNet Blocks for Connection} 
The timestep and audio features are processed through a linear layer to produce Scale and Shift embeddings, which are then applied to the combined input generated by the Reference-Guided Priors module. By applying the Scale and Shift transformations, a link is established between audio-driven cues and spatiotemporal keypoint adjustments. This transformed input is subsequently passed through the Spatiotemporal-Aware Attention Network via a residual connection, preserving essential characteristics from the original integrated input.

\bmhead{Spatiotemporal-Aware Self-Attention}
Spatiotemporal-Aware Attention Network integrates a self-attention mechanism that ensures coherence across spatial and temporal dimensions, aligning predicted motion keypoints with audio cues while capturing dependencies across frames. This mechanism introduces a spatiotemporal bias to emphasize both recent frame transitions and spatial consistency within each frame, ensuring fluid motion progression. Intermediate features from ResNet blocks are projected into query, key, and value vectors ($Q, K, V$), which are further enriched with spatial-temporal positional encodings to capture dynamic, time-sensitive patterns. The resulting embeddings enable the model to interpret relationships across time points and spatial regions in sync with audio rhythm. Attention weights are derived by computing the dot product of the encoded $Q$ and $K$ vectors, followed by softmax normalization, and applied to the $V$ vectors in a weighted summation. Residual connections preserve original feature information, yielding refined motion keypoints aligned with the temporal structure and spatial details driven by audio. We verified the importance of this attention mechanism in ensuring lip synchronization and video smoothness in ablation experiments.

\subsection{Face Render}
\label{sec:face render}
The face rendering component synthesizes the final talking head animation by using motion keypoints and transformation parameters predicted by the Keypoint-based Spatiotemporal Diffusion model, along with appearance features from the reference image. These appearance features, encapsulating the subject's identity and texture, maintain the visual fidelity of the animated face. Meanwhile, the motion keypoints drive audio-synchronized facial expressions, lip movements, and head poses. Rendering is achieved through the LivePortrait~\cite{guo2024liveportrait} warping and decoder modules, which deform the facial structure and reconstruct the visual details frame-by-frame to produce high-quality, synchronized video aligned with the input audio.

\section{Experiments}\label{sec4}
\subsection{Experiments Setup}
\bmhead{Datasets}
The proposed model were trained on the VoxCeleb dataset~\cite{Nagrani19}, which includes over 100,000 video clips of 1,251 subjects. Due to misalignments in some video and audio pairs within VoxCeleb, 4,282 aligned video-audio pairs were selected in the training stage. Following the image animation method~\cite{siarohin2019first}, we cropped the original videos and resized them to 256×256. Audio inputs were downsampled to 16k Hz and converted to the mel-spectrogram features, following the settings in Wav2lip~\cite{prajwal2020lip}. For evaluation, the first 8 seconds of 349 videos from the HDTF dataset~\cite{zhang2021flow} were used, with the first frame of each video serving as the reference image for video generation.

\bmhead{Evaluation Metric} 
The effectiveness of our method was demonstrated through several widely recognized evaluation metrics. For lip synchronization and mouth shape accuracy, we used perceptual metrics from Wav2Lip~\cite{prajwal2020lip}, specifically the distance score (LSE-D) and confidence score (LSE-C). To analyze head motion, we assessed the diversity of generated motions by calculating the standard deviation of head motion feature embeddings extracted from the generated frames using Hopenet~\cite{ruiz2018fine}. Additionally, we evaluated the quality of generated frames with Frechet Inception Distance (FID)~\cite{heusel2017gans} for frame realism and cumulative probability blur detection (CPBD)~\cite{narvekar2011no} for image sharpness. Identity preservation was assessed by computing the cosine similarity (CSIM) between identity embeddings of source images and generated frames, using ArcFace~\cite{deng2019arcface} for embedding extraction. Finally, we measured inference efficiency by reporting Inference Time in frames per second (FPS), indicating the number of frames generated per second.

\bmhead{Implementation Details}
All experiments were conducted on a single NVIDIA GeForce RTX 4090.
The proposed KDTalker is an end-to-end model that requires only a single training session, and the keypoint diffusion model consists of 42.93M parameters.
Furthermore, The diffusion timesteps are set to 1,000 during the training phase, while DDIM~\cite{song2021ddim} is employed for faster inference with 50 steps. 
The model processed 64 audio frames per pass, with a generated image resolution of 512x512. The experiments utilized the AdamW optimizer~\cite{kingma2014adam} with a learning rate schedule that includes Warmup and cosine annealing, where the rate initially increases linearly to 5.12$e^{-4}$ before decaying. 
Additionally, the batch size for experiments was set to 256. Finally, the unsupervised motion keypoints and parameters in our training data were extracted using the pre-trained video-driven method LivePortrait~\cite{guo2024liveportrait}.

\subsection{Comparison with the SOTA Methods}
The proposed method was compared with several state-of-the-art approaches for audio-driven talking portrait generation, including keypoint-based methods like SadTalker~\cite{zhang2023sadtalker} and Real3DPortrait~\cite{yereal3d}, as well as image-based approaches like AniTalker~\cite{liu2024anitalker} and AniPortrait~\cite{wei2024aniportrait}. All the methods are compared quantitatively and qualitatively.

\bmhead{Quantitative Comparison}
As shown in Table~\ref{tab:1}, the proposed method outperforms existing methods in lip synchronization, head motion diversity, video quality, and inference speed. Specifically, it achieves the highest LSE-C and lowest LSE-D, demonstrating superior lip-sync accuracy, closely matching real video performance. In head motion, our method demonstrates the highest diversity attributed to its powerful generative capabilities with unsupervised keypoints. Though our method may generate an even higher diversity than that of real video, it leads to the smallest diversity difference. For video quality, our CPBD achieves a performance comparable to SadTalker, while AniPortrait achieves the highest CPBD thanks to Stable Diffusion~\cite{rombach2022high} but has the slowest inference speed. Nevertheless, our best FID and CSIM scores highlight the superior visual quality and structural accuracy of our output. Furthermore, our inference speed reaches 21.678 FPS, demonstrating its suitability for real-time applications. These results demonstrate our model's superiority over state-of-the-art methods.

\begin{table*}[ht]
\caption{Quantitative comparison with the state-of-the-art methods on HDTF dataset.}
\centering
\resizebox{\textwidth}{!}{%
\begin{tabular}{l|l|cc|c|ccc|c}
\toprule
\multirow{2}{*}{\centering Method} & \multirow{2}{*}{\centering Venue} & \multicolumn{2}{c|}{Lip Synchronization} & Head Motion & \multicolumn{3}{c|}{Video Quality} & Inference Time \\ \cline{3-9}  
               & & LSE-C $\uparrow$ & LSE-D $\downarrow$ & Diversity $\uparrow$ & FID $\downarrow$ & CPBD $\uparrow$ & CSIM $\uparrow$ & FPS $\uparrow$ \\ \hline
Real Video    & - & 8.243 & 6.929  & 0.639     & 0.000  & 0.353 & 1.000 & 25\\ \hline
SadTalker~\cite{zhang2023sadtalker} & CVPR'23  & 7.121 & 7.813  & 0.494     & 12.701 & 0.280 & 0.929 & 21.277\\
Real3DPortrait~\cite{yereal3d} & ICLR'24 & 7.004 & 7.804  & 0.118     & 17.428 & 0.239 & 0.937 & 3.092\\
AniTalker~\cite{liu2024anitalker} &  MM'24 & 7.107 & 8.077  & 0.374     & 10.244 & 0.305 & 0.893 & 14.045\\
AniPortrait~\cite{wei2024aniportrait} & ArXiv'24 & 3.438 & 10.946 & 0.353     & 14.480 & \textbf{0.401} & 0.928 & 1.555\\ 
\rowcolor{gray!20}
Ours & - & \textbf{7.326} & \textbf{7.548}  & \textbf{0.760}     & \textbf{9.756}  & 0.277 & \textbf{0.949} & \textbf{21.678} \\ \bottomrule
\end{tabular}%
}
\label{tab:1}
\end{table*}

\begin{figure}[t]
\centering
   \includegraphics[width=1\linewidth]{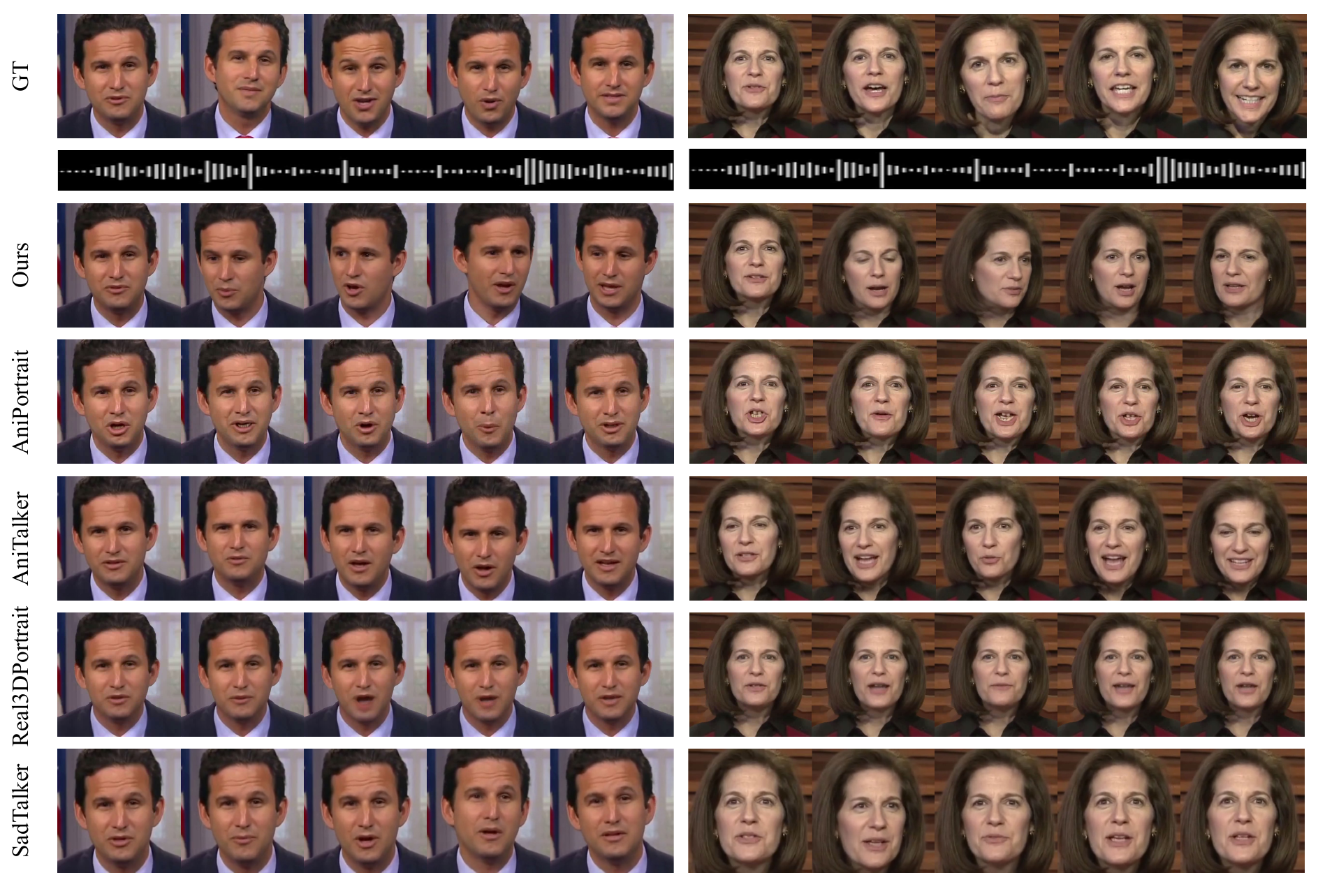}
   \caption{Qualitative comparison with the state-of-the-art methods on HDTF dataset.}
   \label{fig:qualitative comparison}
\end{figure}

\begin{figure}[t]
  \centering
   \includegraphics[width=1\linewidth]{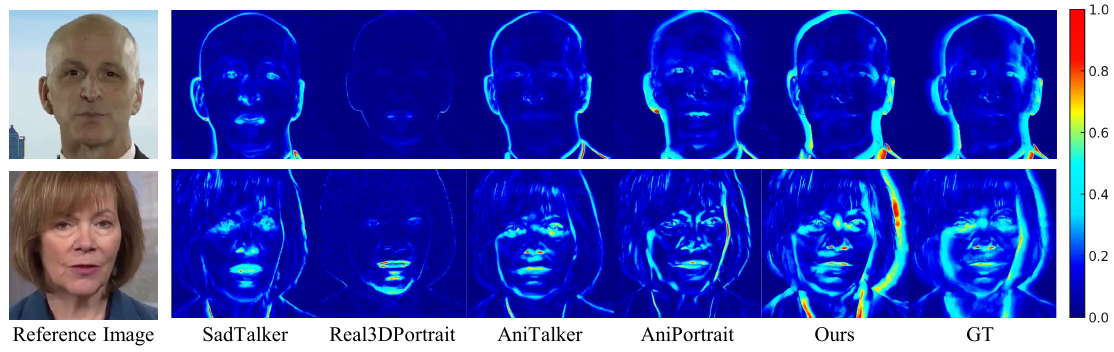}
   \caption{Qualitative comparison of head motion diversity between existing methods and our proposed approach.}
   \label{fig:head motion}
\end{figure}

\bmhead{Qualitative Comparison} 
Qualitative visual results of various methods are provided, as shown in Fig.~\ref{fig:qualitative comparison}. In addition, we also provide the GT results under the corresponding audio frames to visualize the lip synchronization of our method. As can be seen from the figure, our method has lip movements similar to the GT video, but has richer head movements. The head movement range of all the competitors is relatively small and lacks diversity. To further verify the head diversity of our method, the motion heatmaps of various methods were calculated and shown in Fig.~\ref{fig:head motion}. In addition, AniPortrait's lips are distorted, resulting in poor lip synchronization performance. Please refer to our supplementary videos for a clearer comparison.

\subsection{User Studies}
A user study was conducted to evaluate the performance of all methods.
We selected 20 sample images and corresponding audio from the HDTF dataset as tests. These reference sample images contain different genders, ages, expressions, poses, and backgrounds to show the robustness of our method. The study involved 20 participants, who were asked to subjectively evaluate and select the best-performing method for each sample on each criterion, which included: lip synchronization, the diversity of head motion and overall naturalness. In particular, the participants comprised 13 males and 7 females with diverse professional backgrounds, including doctors, algorithm researchers, product managers, engineers, quality inspectors, and students of different grades. To eliminate any potential bias caused by the order of presentation, the sequence of methods in the comparative videos was randomized for each sample, ensuring the authenticity and fairness of the evaluations.

The results are shown in Table~\ref{tab:2}. In terms of the three dimensions, participants prefer our generated videos. It is worth noting that the overall naturalness scores of 0\% for AniPortrait~\cite{wei2024aniportrait} and Real3DPortrait~\cite{yereal3d} can be attributed to different issues. AniPortrait~\cite{wei2024aniportrait} has obvious lip distortion, while Real3DPortrait~\cite{yereal3d} suffers from fixed poses. More user study details can be seen in the supplemental material.

\begin{table*}[h]
\caption{User study showing the percentage of participants who selected a method as the best across three evaluation dimensions.}
\centering
\resizebox{0.6\textwidth}{!}{%
\begin{tabular}{l|c|c|c}
\toprule
\multirow{2}{*}{\centering Method} & Lip  & Head & Overall \\
               & Sync.  & Diversity  & Naturalness\\ \hline
SadTalker~\cite{zhang2023sadtalker}       & 14.5\% & 22.8\%  & 15.8\%\\
Real3DPortrait~\cite{yereal3d}            & 3.8\% & 0.5\%  & 0.0\%\\
AniTalker~\cite{liu2024anitalker}         & 39.0\% & 13.0\%  & 38.5\% \\
AniPortrait~\cite{wei2024aniportrait}     & 2.0\% & 2.0\%  & 0.0\% \\
\rowcolor{gray!20} 
Ours                                      & \textbf{40.7\%} & \textbf{61.7\%}  & \textbf{45.7\%} \\ \bottomrule
\end{tabular}%
}
\label{tab:2}
\end{table*}

\subsection{Analytical Experiments}
\subsubsection{Ablation Study for Main Components} 
First, we verify the important role of the typical key point $x_{c}$ in the reference image. As shown in Table~\ref{tab:3}, removing $x_{c}$ reduces lip synchronization by approximately 8\%, as $x_{c}$ offers crucial prior keypoint information.

The Spatiotemporal-Aware Attention module captures temporal dependencies between frames by modeling their sequential relationships through Rotary Position Embedding, thereby establishing proximity relationships between frames. This design enables the model to simultaneously address short-term dynamics and long-term temporal evolution, rather than treating all video frames as independent inputs. Excluding the spatiotemporal attention mechanism causes a sharp drop in lip synchronization, resulting in excessive and incoherent head movements that lead to abnormally high head motion diversity. We observed that lip synchronization is stable for frontal head poses but declines significantly with larger head movements (e.g., profiles or side views). To validate the advantages of RoPE, we conducted additional ablation studies, as shown in Table~\ref{tab:3}. The results show that RoPE significantly enhances lip synchronization (higher LSE-C, lower LSE-D) and head motion diversity compared to vanilla self-attention.

Overall, our method effectively balances coherent and naturally diverse head movements with strong lip synchronization, demonstrating the advantages of incorporating RoPE and spatiotemporal attention in generating high-quality talking head videos.


\begin{table*}[h]
\caption{Ablation study for main components in proposed method.}
\centering
\resizebox{0.6\textwidth}{!}{%
\begin{tabular}{l|cc|c}
\toprule
\multirow{2}{*}{\centering Method} & \multicolumn{2}{c|}{Lip Synchronization} & Head Motion \\ \cline{2-4} 
        & LSE-C $\uparrow$ & LSE-D $\downarrow$ & Diversity $\uparrow$ \\ \hline
w/o Reference $x_{c}$       & 6.656 & 8.203  & 0.791  \\
w/o Attention   & 1.360 & 11.777  & \textbf{2.292}  \\
w/o RoPE        & 7.225 & 7.663   & 0.706 \\
\rowcolor{gray!20}
Ours Full      & \textbf{7.326} & \textbf{7.548}  & 0.760 \\ \bottomrule
\end{tabular}%
}
\label{tab:3}
\end{table*}

\subsubsection{Comparison of Generative Methods}  
To demonstrate the advantages of diffusion models, we compared our model with versions that replace the generator with VAE~\cite{kingma2013vae} and GAN~\cite{goodfellow2020gan}, keeping the overall architecture unchanged.
As shown in Table~\ref{tab:4}, the VAE model achieves diverse head movements but lacks accurate lip-audio synchronization, while GAN suffers from convergence issues, leading to distorted outputs.
The combined VAE+GAN approach, similar to SadTalker, shows some audio-to-lip mapping but fails to maintain temporal consistency in head movements, resulting in unconvincing outputs.
In contrast, the diffusion-based method achieves superior synchronization and coherent, naturally diverse head poses, demonstrating its effectiveness for high-quality talking portrait generation.

\begin{table*}[h]
\caption{Comparison of our model with different generators}
\centering
\setlength {\abovecaptionskip} {.2cm}
\setlength {\belowcaptionskip} {0.cm}
\resizebox{0.65\textwidth}{!}{%
\begin{tabular}{l|cc|c}
\toprule
\multirow{2}{*}{\centering Method} & \multicolumn{2}{c|}{Lip Synchronization} & Head Motion \\ \cline{2-4} 
        & LSE-C $\uparrow$ & LSE-D $\downarrow$ & Diversity $\uparrow$ \\ \hline
VAE~\cite{kingma2013vae}     & 0.665 & 13.879  & 0.846  \\
GAN~\cite{goodfellow2020gan}   & - & -  & -  \\
VAE~\cite{kingma2013vae} + GAN~\cite{goodfellow2020gan}    & 1.232 & 13.284  & \textbf{0.876}  \\
\rowcolor{gray!20}
Diffusion (Ours) & \textbf{7.326} & \textbf{7.548}  & 0.760 \\ \bottomrule
\end{tabular}%
}
\label{tab:4}
\end{table*}

\subsubsection{Ablation of Face Render}
High-quality face render is crucial for generating realistic and engaging animated videos, as it directly affects both the visual authenticity and the coherence of facial movements.
We compared Face-vid2vid~\cite{wang2021facevid2vid} and LivePortrait~\cite{guo2024liveportrait} across key dimensions such as lip synchronization and overall video quality to evaluate the effectiveness of different face rendering methods.
As shown in Table~\ref{tab:5}, LivePortrait demonstrates distinct advantages in capturing accurate lip movements and enhancing visual authenticity through a greater number of unsupervised keypoints, which allow it to represent facial details more precisely.
Although Face-vid2vid performs slightly better in certain video quality metrics like image clarity, LivePortrait’s marginally lower scores do not significantly impact its overall output.
The slight trade-off in sharpness is offset by LivePortrait’s superior lip synchronization and structural realism, making it highly effective for producing lifelike animated portraits.

\begin{table*}[h]
\caption{Ablation of Face Render. F.V. denotes Face-vid2vid, while L.P. stands for LivePortrait, which is used in our method.}
\centering
\resizebox{0.7\textwidth}{!}{%
\setlength\tabcolsep{2 pt}
\begin{tabular}{l|c|cc|ccc}
\toprule
\multirow{2}{*}{\centering Method} & \multirow{2}{*}{\centering Points} & \multicolumn{2}{c|}{Lip Synchronization} & \multicolumn{3}{c}{Video Quality} \\ \cline{3-7} 
       & & LSE-C $\uparrow$ & LSE-D $\downarrow$ & FID $\downarrow$ & CPBD $\uparrow$ & CSIM $\uparrow$ \\ \hline
F.V.~\cite{wang2021facevid2vid}    & 15 & 5.579 & 9.228 & \textbf{8.625} & \textbf{0.307} & 0.940\\ 
\rowcolor{gray!20}
L.P.~\cite{guo2024liveportrait}    & \textbf{21} & \textbf{7.326} & \textbf{7.548} & 9.756  & 0.277 & \textbf{0.949} \\ \bottomrule
\end{tabular}%
}
\label{tab:5}
\end{table*}

\subsubsection{Ablation of Inference Step}
To balance performance and efficiency during inference, we conducted an ablation study to explore the impact of different sampling steps when using the DDIM strategy. Table \ref{tab:Ablation of inference step} shows the results, focusing on lip synchronization and video quality. The findings show that increasing the number of steps initially improves both metrics, particularly from 1 to 5 steps, but further increases yield diminishing returns. The best video quality, reflected in the lowest FID, is observed at 50 steps, while CPBD and CSIM remain consistently high. Although fewer steps slightly enhance lip synchronization, 50 steps provide the optimal balance between quality, synchronization, and computational cost. This motivated our decision to use 50 steps for inference to achieve efficient and high-quality results.

\begin{table*}[h]
\caption{Ablation of inference step}
\centering
\resizebox{0.60\textwidth}{!}{%
\begin{tabular}{c|cc|c|c|c}
\toprule
\multirow{2}{*}{\centering Step} & \multicolumn{2}{c|}{Lip Synchronization} & \multicolumn{3}{c}{Video Quality} \\ \cline{2-6} 
        & LSE-C $\uparrow$ & LSE-D $\downarrow$ & FID $\downarrow$ & CPBD $\uparrow$ & CSIM $\uparrow$ \\ \hline
1       & 0.817 & 12.630 &  19.058 & 0.262 & 0.823\\
5      & \textbf{7.455} & \textbf{7.424} & 10.226 & 0.277 & 0.949\\
10      & 7.448 & 7.436 & 9.939 & \textbf{0.278} & 0.948 \\
20      & 7.394 & 7.501 & 9.797 & 0.277 & 0.945 \\
\rowcolor{gray!20}
50      & 7.326 & 7.548 & \textbf{9.756} & 0.277 & \textbf{0.949} \\
100      & 7.264 & 7.614 & 9.987 & 0.277 & 0.948\\
200      & 7.221 & 7.633 & 9.901 & 0.277 & 0.948 \\ \bottomrule
\end{tabular}%
} 
\label{tab:Ablation of inference step}
\end{table*}

\subsubsection{Ablation of Frame Number}
To improve video coherence, particularly in lip synchronization and head motion, we explored the impact of different frame numbers. Table \ref{tab:Ablation of frame number} shows that increasing the number of frames generally improves lip synchronization and head motion diversity, with 64 frames achieving the best performance. Using 64 frames allows for better temporal dependencies between frames, which helps achieve smoother transitions and more natural motion. This results in improved lip synchronization and head motion consistency, making 64 frames a favourable choice for enhancing video quality and continuity.

\begin{table*}[h]
\caption{Ablation of frame number.}
\centering
\resizebox{0.60\textwidth}{!}{%
\begin{tabular}{c|cc|c}
\toprule
\multirow{2}{*}{\centering Frame Number} & \multicolumn{2}{c|}{Lip Synchronization} & Head Motion \\ \cline{2-4} 
        & LSE-C $\uparrow$ & LSE-D $\downarrow$ & Diversity $\uparrow$ \\ \hline
8       & 6.875 & 7.928  & 0.673 \\
16      & 6.912 & 7.904  & 0.687 \\
32      & 7.256 & 7.636  & 0.686 \\
\rowcolor{gray!20}
64      & \textbf{7.326} & \textbf{7.548}  & \textbf{0.760}  \\ \bottomrule
\end{tabular}%
}
\label{tab:Ablation of frame number}
\end{table*}

\section{Conclusion}\label{sec5}
In this work, we presented KDTalker, a novel framework for generating natural and dynamic talking portraits that effectively combines the strengths of unsupervised 3D keypoint-driven animation with the diversity capabilities of diffusion models. KDTalker addresses the limitations of previous keypoint-based and image-based methods by offering both high lip synchronization accuracy and rich head pose diversity, which are essential for producing realistic and expressive digital portraits. 
Unlike traditional 3DMM with fixed keypoints, KDTalker adapts to varying facial information densities, capturing nuanced expressions. It also integrates a spatiotemporal attention mechanism to capture long-range audio-keypoint dependencies, ensuring precise lip synchronization and coherent head poses that reflect speech dynamics.
Through extensive experiments, KDTalker achieves state-of-the-art performance on multiple metrics, becoming a promising solution for real-time applications.

\bmhead{Limitation}
The KDTalker framework, while innovative, has limitations primarily arising from its reliance on keypoint detection and transformation. Since KDTalker relies on precise 3D keypoint detection for facial rendering, noisy data or complex facial features can result in misalignments and distortions in the animation. Additionally, the framework struggles with occlusions, where parts of the face are blocked by external objects. These occlusions can disrupt the keypoint transformation process, resulting in artefacts such as blurred edges or distorted features, particularly in areas crucial to facial expressions like the eyes, mouth, and nose. These limitations underscore the framework's dependence on precise keypoint detection, which remains vulnerable to errors in complex scenarios, impacting both the accuracy and expressiveness of the generated animations.

\bibliography{sn-bibliography}

\end{document}